\documentclass[conference]{IEEEtran}
\IEEEoverridecommandlockouts
\usepackage{cite}
\usepackage{amsmath,amssymb,amsfonts}
\usepackage{algorithmic}
\usepackage{graphicx}
\usepackage{textcomp}
\usepackage{xcolor}
\def\BibTeX{{\rm B\kern-.05em{\sc i\kern-.025em b}\kern-.08em
    T\kern-.1667em\lower.7ex\hbox{E}\kern-.125emX}}
\usepackage{balance}

\usepackage{hyperref}

\begin{document}

\title{Towards Context-Aware Human-like Pointing Gestures with RL Motion Imitation 
\\

\thanks{* equal contributions. This work was financed by Digital Futures, project ``Advanced Adaptive Intelligent Systems" and Swedish Research Council, grant nr 2018-05409.}
}

\author{\IEEEauthorblockN{Anna Deichler*}
\IEEEauthorblockA{\textit{Division of Speech, Music and Hearing} \\
\textit{KTH Royal Institute of Technology}\\
Stockholm, Sweden \\
deichler@kth.se }
\and
\IEEEauthorblockN{Siyang Wang*}
\IEEEauthorblockA{\textit{Division of Speech, Music and Hearing} \\
\textit{KTH Royal Institute of Technology}\\
Stockholm, Sweden \\
siyangw@kth.se } \\

\IEEEauthorblockN{Jonas Beskow}
\IEEEauthorblockA{\textit{Division of Speech, Music and Hearing} \\
\textit{KTH Royal Institute of Technology}\\
Stockholm, Sweden \\
beskow@kth.se }
\and
\IEEEauthorblockN{Simon Alexanderson}
\IEEEauthorblockA{\textit{Division of Speech, Music and Hearing} \\
\textit{KTH Royal Institute of Technology}\\
Stockholm, Sweden \\
simonal@kth.se } 

}

\maketitle

\begin{abstract}
Pointing is an important mode of interaction with robots. While large amounts of prior studies focus on recognition of human pointing, there is a lack of investigation into generating context-aware human-like pointing gestures, a shortcoming we hope to address. We first collect a rich dataset of human pointing gestures and corresponding pointing target locations with accurate motion capture. Analysis of the dataset shows that it contains various pointing styles, handedness, and well-distributed target positions in surrounding 3D space in both single-target pointing scenario and two-target point-and-place. We then train reinforcement learning (RL) control policies in physically realistic simulation to imitate the pointing motion in the dataset while maximizing pointing precision reward. We show that our RL motion imitation setup allows models to learn human-like pointing dynamics while maximizing task reward (pointing precision). This is promising for incorporating additional context in the form of task reward to enable flexible context-aware pointing behaviors in physically realistic environment while retaining human-likeness in pointing motion dynamics. 
\end{abstract}


\section{Introduction}
There is an increasing number of studies on human-robot interaction using pointing gestures \cite{tolgyessy2017foundations,abidi2013human,haring2012studies}. The emphasis has so far been put on recognizing pointing from human user and distinguishing the pointed object, while generation of pointing behavior in robots is less studied. Pointing as a functional gesture has two important aspects; legibility, or how accurate is the pointing \cite{holladay2014legible}, and motion dynamics, or the human-likeness of the pointing motion itself. The latter aspect is especially lacking in prior studies. Low human-likeness in pointing motion can hinder human perception of the robot intention. Moreover, humans point with varying dynamics to convey extra information besides just what is being pointed at. For example, a quick point and stop motion when referring to an object shows certainty, while a slower pointing with sub-movements (e.g. hovering around pointed object) indicates less confidence or certainty. The dynamics of pointing motion can also be interpreted as style factors that may convey information about individual personality.


%

In recent years, methods for generating naturalistic and believable gesture for virtual characters and social robots have advanced tremendously, thanks to new methods in probabilistic motion modelling using deep architectures and the availability of extensive motion capture datasets \cite{alexanderson2020style,yoon2020speech}. However, such methods require a large amount of data (motion capture) and are not designed to adapt to unseen environments. 



In this paper we propose to use a combination of data-driven and physically based reinforcement learning (RL) for communicative gestures. The reinforcement learning component allows the model to flexibly adapt itself to task rewards (e.g. pointing accuracy) during simulation training or real-world adaptation. We take inspiration from prior work in the domain of physically based character animation \cite{peng2018deepmimic,peng2021amp}, and discuss how these approaches may be applied to gestures that are communicative in nature and also dependent on the context of the environment. Our initial experiments show promise of RL-based motion imitation approach for generating dynamic, accurate, and human-like pointing gestures.

We also collect a new motion-captured pointing gesture dataset. We present summary statistics and visualization of target positions, handedness (left or right), and range of movement in the dataset. They show that the dataset contains well-distributed target positions and diverse range of movement, thus it has the potential to be used in several different tasks, such as generating pointing behaviors (this paper), or training a model to identify pointing gesture and pointed target position. We show that the arm of the actor is not precisely aligned with target position. This suggests that a simple metric like the alignment angle is \textit{not} enough to, (1) be the only RL reward in generating human-like pointing as our data analysis show that humans do not point "accurately" in terms of minimizing alignment angle, (2) be the only criteria in discriminating the object a user is pointing to.




\section{Related work}




Pointing has been extensively studied in human-robot interaction. \cite{7333604} has developed a a human-robot gesture language for two-way communication in  pick-and-place  tasks, which included pointing. Prior studies have concentrated on the recognition of human pointing gestures  \cite{NICKEL20071875},\cite{6281384}. There are fewer studies on generation of pointing behavior in robots. \cite{holladay2014legible} proposes a mathematical optimization approach to make legible pointing, i.e. disambiguation of closely situated objects. A more recent work \cite{zhao2020actor} proposes a reinforcement learning approach with actor-critic setting to achieve legible pointing. An imperial study on making humanoid robot exhibit guide \cite{iio2019human} employs a pointing generation sub-system. These studies mostly focus on pointing accuracy, and do not specifically address human-likeness of pointing behaviors from a motion dynamics perspective. 

In human communication, deictic gestures (pointing) are ubiquitous, and one of the first means of communication developed in infancy \cite{kita2003pointing}. Pointing helps us communicate more effectively and robustly, guide visual attention \cite{doi:10.1177/1461445605054404}, and understand other people's intentions \cite{Camaioni2004TheRO}. Pointing typically convey complementary information to the accompanying speech rather than redundant \cite{de2016deictic}. Studies on human pointing have also revealed that humans are able to point accurately with variable speed \cite{adamovich1994influence} and to point with variable trajectories \cite{blinch2018trajectory}. This presents challenge for learning physically realistic and human-like pointing gestures in robots.

Gesture synthesis for virtual agents and social robots has recently shifted from rule-based systems employing instructions given in a markup language \cite{van2014asaprealizer, kopp2006towards} to data-driven methods generating gestures directly from speech acoustics or text \cite{alexanderson2020style, yoon2020speech}. This development has increased naturalness and ease-of-use to the trade-off of losing ability to generate complementary gestures. This paper aims to bridge the gap by proposing a learning environment and framework to learn complementary aspects of speech and pointing gestures.

Reinforcement learning (RL) has been extensively used in robotics studies, since compared to other machine learning methods it allows the agent to learn from interacting with its environment, therefore it allows adaptability, compared to passive supervised machine learning settings \cite{RLarticle}. In recent years, accurate physics engines enabled training RL agents in ever more realistic scenarios. However, it is challenging to produce natural motor behavior for physically simulated humanoids. Therefore recent works have explored a combination of motion capture based imitation learning with RL to achieve generalizable and human-like skill learning \cite{peng2018deepmimic},\cite{peng2021amp},\cite{pmlr-v119-hasenclever20a}.





\section{Motion Capture Data}




A pointing gesture dataset was recorded in a optical motion capture studio equipped with 16 Optitrack Prime 41 cameras. The actor wore a suit with 50 passive markers, a pair of Manus data gloves (hand motion capture) and a head-mounted iPhone 12 (face motion and voice capture). The motion capture system recorded the actor's joint positions, as well as the target positions in 3D space. 
Three different pointing tasks were executed: single target pointing, two targets selection and two targets moving (point-and-place). In each setting the targets were moved around in order to get a coverage of the surrounding space, while the actor was in a stationary position. Between each pointing task the actor returned to a neutral stance. The beginning and the end of the movement is defined by the pointing hand leaving and returning to the initial, downward position.


The dataset also includes recorded speech as the actor is pointing. Example utterances are, "I want that one" (in single-target pointing), and "Put this one there" (in two-target pointing and placing). The speech data is synchronized with mocap. This provides extra context for training models to generate pointing behaviour, and recognition of pointing timing, perhaps using speech as extra input to better identify the instant of pointing instead of relying on the motion itself. 

\subsection{Data analysis}
We focus our data analysis on the single target pointing task, since models in this study are trained on single target clips.  The single target dataset is parsed into single arm pointing movement based on peaks in displacement on the sagittal plane. The resulting target position counts can be read in Table \ref{tab1:target_counts}. The total sum of single target positions in the dataset is 83. There are 32 left-handed and 51 right-handed pointing movements. The actor's dominant hand is right hand. 

\begin{figure}[!t]
  \centering

  \includegraphics[width=0.9\linewidth]{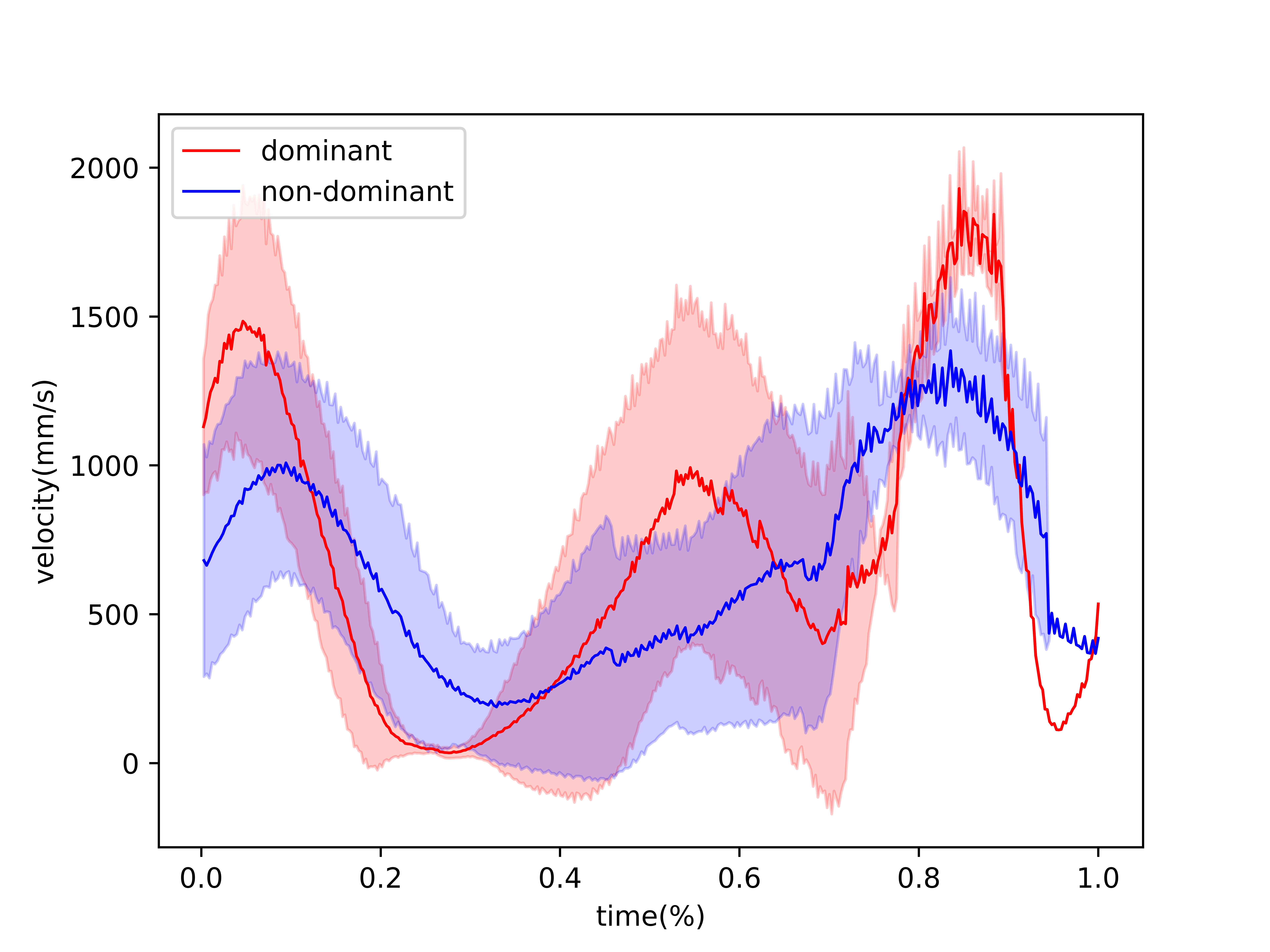}
  \\

  \includegraphics[width=0.9\linewidth]{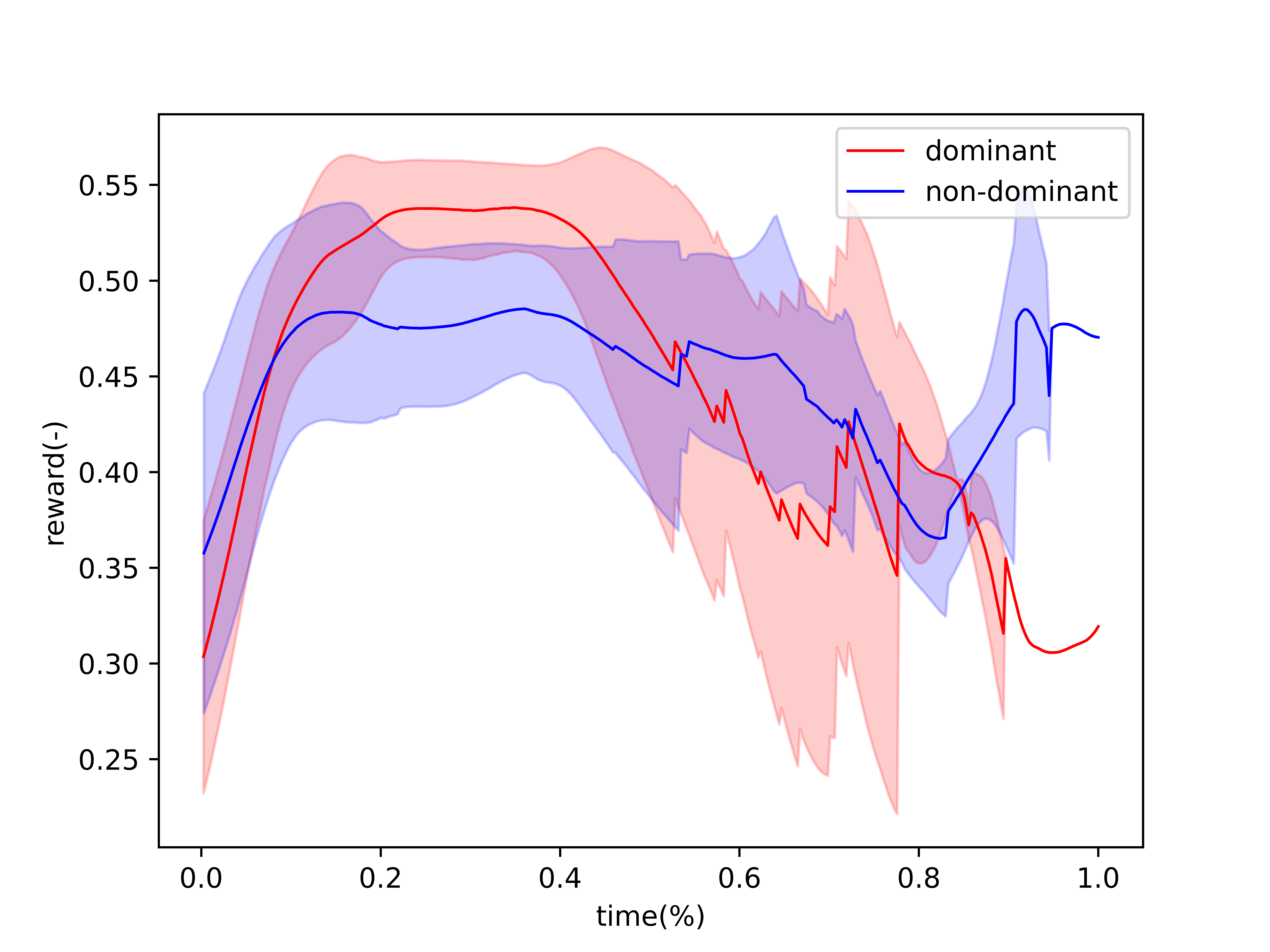}
  \\
  \caption{Average hand velocity profiles and average pointing precision profiles  from  the single pointing task for dominant and non-dominant hands.}
  \label{fig:data_analysis}
\end{figure}

We analyse the pointing movements in terms of duration of pointing movement, duration of rise time, peak velocity, average velocity and endpoint precision. The velocity data is obtained from numerical differentiation of the $x,y,z$ positional data from the markers placed on the left and right hands. The average velocity is calculated for the rise phase of the pointing motion. We present the result of our analysis on single target pointing movements in Table \ref{tab1:target_data}. Overall, the table shows that there is clear difference between the dominant and non-dominant hands. The peak and mean velocities of the dominant arm are higher than the non-dominant arm and rise time is also lower. The overall duration is longer for the dominant hand, since the pointing was held out longer.  
 On the top plot in Figure  \ref{fig:data_analysis}  we can see the averaged velocity profiles over the whole pointing movement for both hands. On the bottom plot the averaged pointing precision are depicted for both hands. We measure the pointing precision as a modified pointing angle term proposed in section \ref{ssec:reward}. The maximum measure is 1 and can only be obtained when the arm is fully aligned with the pointed target (angle equals 0), however as seen in the plot, neither dominant or non-dominant hands achieve this measure or come close to achieving it. This shows that human pointing is not accurate in terms of alignment angle an extensively used metric in prior pointing gesture recognition and generation studies. 


\begin{table}[htbp]
\caption{Single target position counts}
\begin{center}
\begin{tabular}{|c|c|c|c|c|}
\hline
\
\textbf{} & \multicolumn{2}{c|}{\textbf{\textit{Front}}} & \multicolumn{2}{c|}{\textbf{\textit{Back}}} \\
\hline
\textbf{} & \textbf{\textit{Left}}& \textbf{\textit{Right}} & \textbf{\textit{Left}} & \textbf{\textit{Right}} \\
\hline
\textbf{\textit{Top}} &  11 & 15 & 8 & 8   \\
\hline
\textbf{\textit{Bottom}} &  14 & 12 & 7 & 8    \\
\hline
\end{tabular}
\label{tab1:target_counts}
\end{center}
\end{table}








\begin{table}[htbp]
\caption{Single target pointing movements}
\begin{center}
\begin{tabular}{|c|c|c|c|}
\hline
\textbf{} & \textbf{\textit{Both}} & \textbf{\textit{Dominant}} & \textbf{\textit{Non-dominant}} \\
\hline
\textbf{\textit{Duration (s)}} &  2.4 $\pm$ 1.05 & 2.61 $\pm$ 1.13 & 2.11 $\pm$ 0.86 \\
\hline
\textbf{\textit{Rise time (s)}} &  0.71 $\pm$ 0.31 & 0.70 $\pm$ 0.30 & 0.73 $\pm$ 0.32 \\
\hline
\textbf{\textit{Peak velocity (mm/s)}} &  1706 $\pm$ 392 & 1825 $\pm$ 384 & 1541 $\pm$ 340 \\
\hline
\textbf{\textit{Mean velocity (mm/s)}} &  849 $\pm$ 302 & 925 $\pm$ 317 & 745 $\pm$ 244 \\
\hline
\multicolumn{4}{l}{$^{\mathrm{a}}$ Unless noted otherwise, all statistics in this paper are reported as [mean] $\pm$ [sd].} \\
\end{tabular}
\label{tab1:target_data}
\end{center}
\end{table}

\begin{figure}[!t]
  \centering
  \bf{front} 
  \vspace{0.5em}
  \\
  \includegraphics[width=1.0\linewidth]{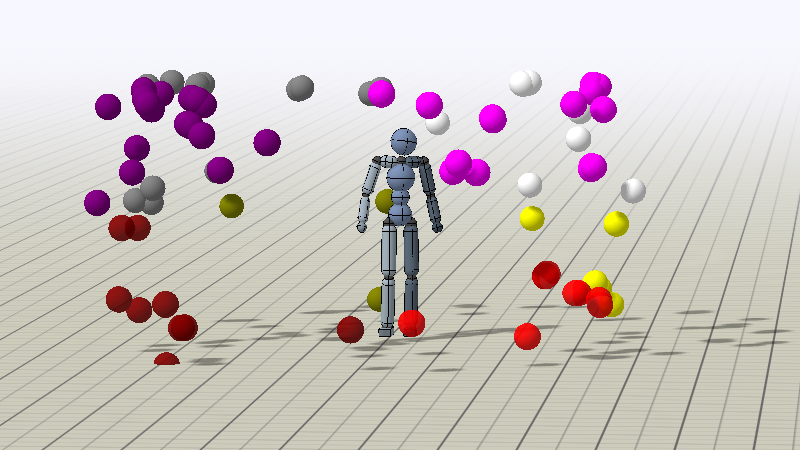}
  \\
  \vspace{0.5em}
  \bf{side} 
  \vspace{0.5em}
  \\
  \includegraphics[width=1.0\linewidth]{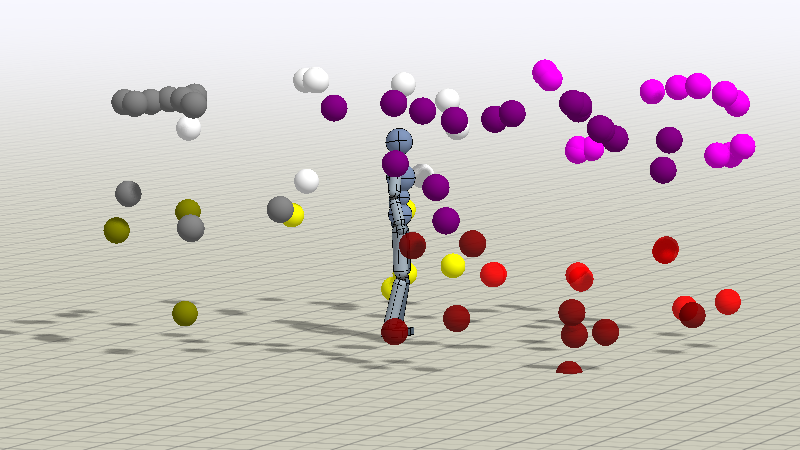}
  \\
  \vspace{0.5em}
  \bf{top} 
  \vspace{1.0em}
  \\
  \includegraphics[width=1.0\linewidth]{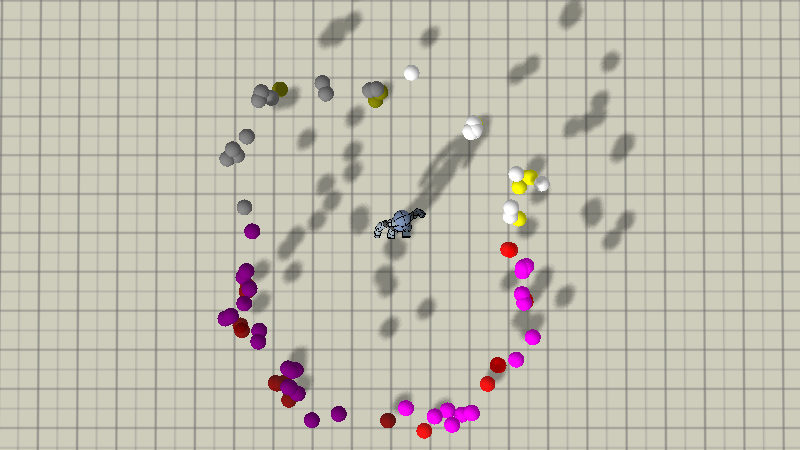}
  \\
  \caption{Target position distribution in single-target pointing.}
  \label{fig:pt_tpos_all}
\end{figure}

\begin{figure}[!t]
  \centering
  \bf{front} 
  \vspace{0.5em}
  \\
  \includegraphics[width=1.0\linewidth]{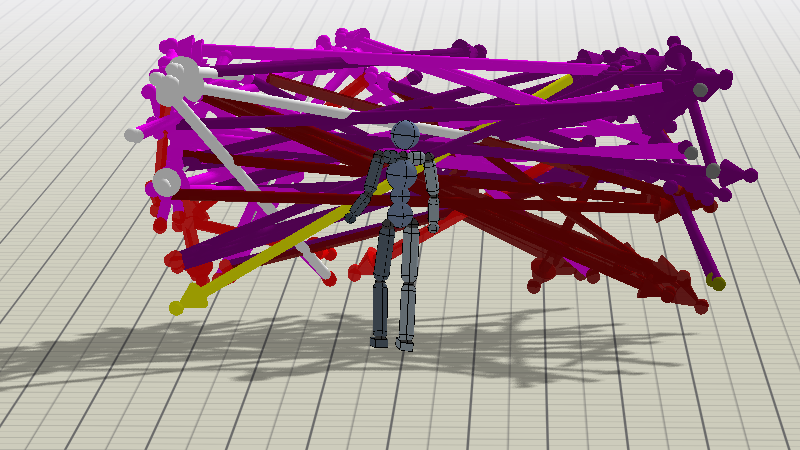}
  \\
  \vspace{0.5em}
  \bf{side} 
  \vspace{0.5em}
  \\
  \includegraphics[width=1.0\linewidth]{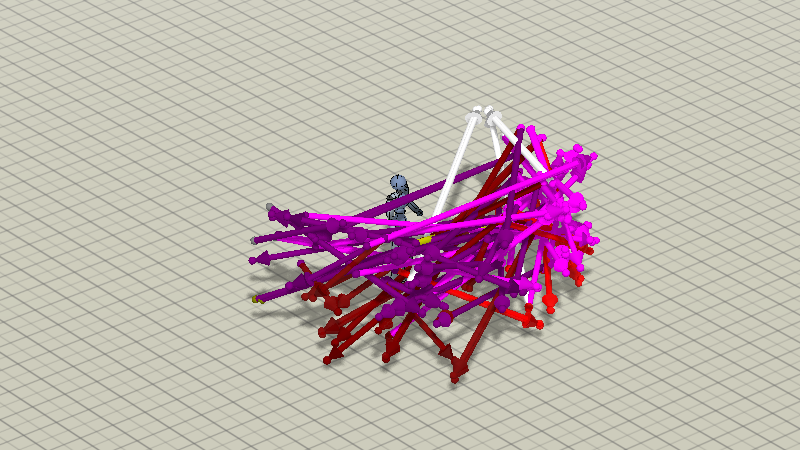}
  \\
  \vspace{0.5em}
  \bf{top} 
  \vspace{1.0em}
  \\
  \includegraphics[width=1.0\linewidth]{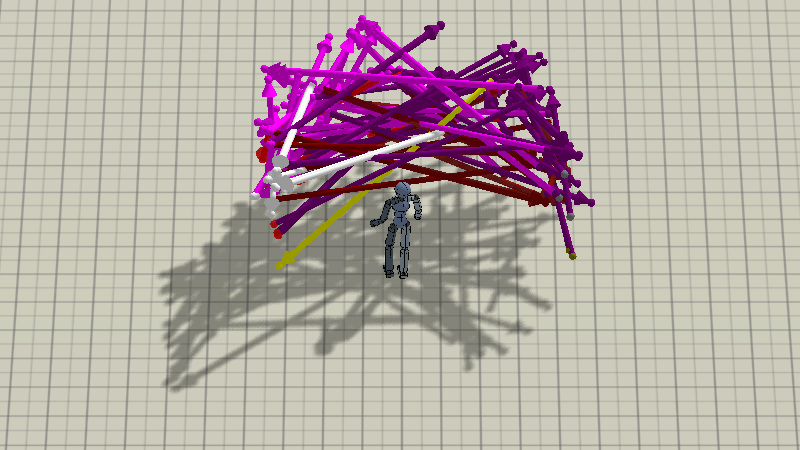}
  \\
  \caption{Target position distribution in two-target pointing and placing. The arrow direction indicating the from- and to- targets. The colors indicating quadrants (same as Figure \ref{fig:pt_tpos_all}) of the end target positions.}
  \label{fig:pt_tpos_all_2}
\end{figure}

We visualize\footnote{All rendering is done using \url{https://github.com/xbpeng/DeepMimic}.} the target distribution in space in Figure \ref{fig:pt_tpos_all} (single-target pointing). Different colors are given to quadrants divided by left/right, top/down, front/back. Positions are distributed evenly in each quadrant. 

Target distribution for two-target pointing task is shown in Figure \ref{fig:pt_tpos_all_2}. The arrows connect two targets that are pointed in a two-target pointing sequence. Our dataset not only covers a wide range of target positions but also a wide range of two-target movement.

\section{Experiment}
Our goal is to achieve human-like pointing in a physically simulated humanoid character that can point to any given target position in its surrounding  3D space. To this end, we use a reinforcement learning (RL) setup, where we aim to learn a control policy which can reproduce the dynamics of the pointing movements from the motion capture dataset and complete the pointing task defined by the pointing reward function. At the same time, the policy should also be robust, namely without falling or other fault behaviors. The RL setup also allows future adaption of arbitrary additional context or modification of rewards to achieve desired style or level of pointing precision.

\subsection{Methodology}
\label{ssec:method}
The learning model is based on Deepmimic (DM) \cite{peng2018deepmimic} and Adversarial Motion Priors (AMP) \cite{peng2021amp} \footnote{Modified implementation from \url{https://github.com/xbpeng/DeepMimic}.}. The model uses motion capture data based imitation and reinforcement learning to achieve human-like motion and to complete a specified task at the same time.

At each time step  $t$ the reward function is the combination of rewards from motion capture imitation $r^{I}_t $ and the task reward $r^{G}_t$ (discussed in section \ref{ssec:reward}), weighted by pre-defined weights $\omega^{I}, \omega^{G}$.
\begin{equation}
r_t=\omega^{I} r^{I}_t + \omega^{G} r^{G}_t 
\end{equation}
For DM models, we keep the original composition of the imitation reward, which is a weighted combination of pose, velocity, end-effector  and center-of-mass rewards 
\begin{equation}
\omega^{I} r^{I}_t =  \omega^{p} r^{p}_t+ \omega^{v} r^{v}_t+ \omega^{e} r^{e}_t+ \omega^{c} r^{c}_t.
\end{equation}
AMP replaces imitation rewards with a single neural-networks-based discriminator. We use same weights for reward terms as original papers\cite{peng2018deepmimic,peng2021amp}. 

We test 4 configurations of DM models. First 3 are generic DM models with pointing target as extra input to the policy network, but they do not receive task reward. These three differ in the number of motion clips used in training, they are denoted as \textbf{DM-4}, \textbf{DM-16}, \textbf{DM-32} where the numbers denote the number of clips. We test an additional DM model \textbf{DM-16-wr} for which the task reward is given along with imitation reward. We test a single AMP model \textbf{AMP-fr}, here \textit{fr} refers to "front raise" because we only give this model front pointing motion clips and all motion clips only contain the arm raising stage of the pointing but not retraction. We use this setup because AMP struggles with distinguishing between the raising and retraction phases of pointing gesture, a problem we hope to solve in future studies. We also restrict target positions for all models to front-facing targets only for easier training and more straightforward comparisons, but the methods can be directly applied to targets in the back.

\subsection{Reward functions for pointing}
\label{ssec:reward}
\begin{figure}[!t]
  \centering

  \includegraphics[width=0.9\linewidth]{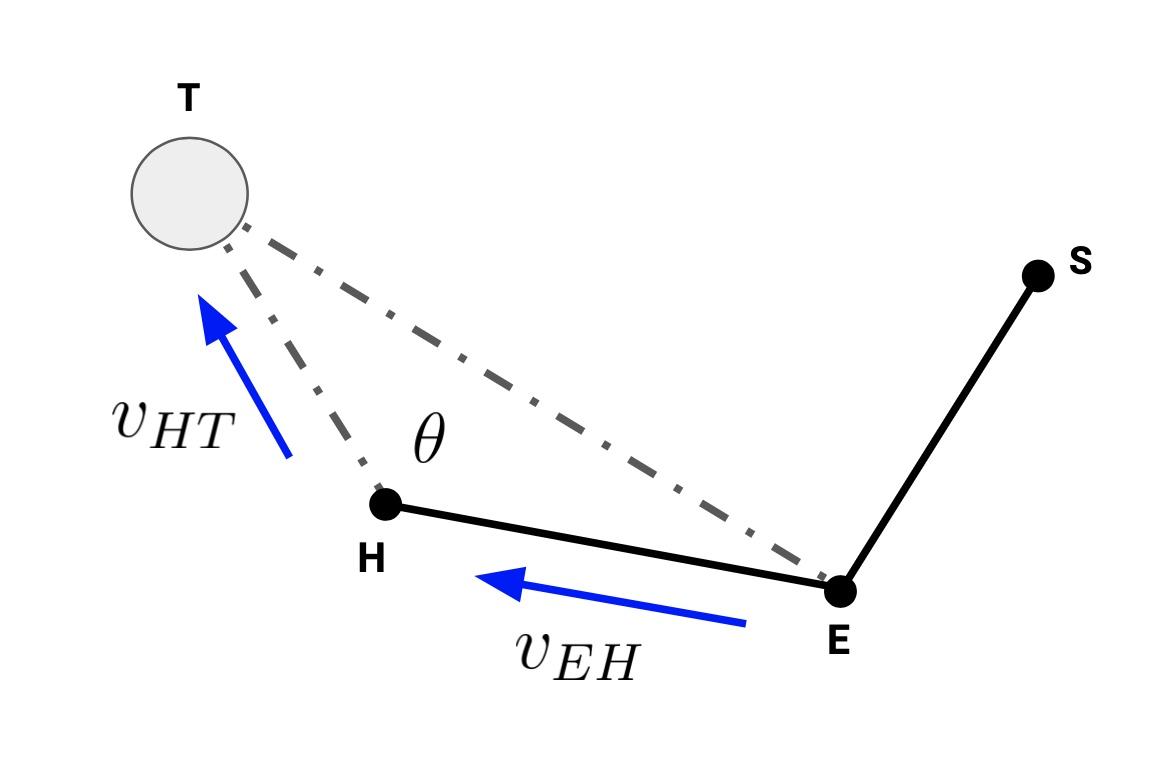}
  \\

  \caption{Pointing reward is defined based on angle between vector from elbow joint (E) to hand joint (H) and vector from hand joint to target position (T) in 3D space.}
  \label{fig:angles}
\end{figure}

In order to achieve the objective of pointing , an appropriate reward function should be devised.  In this paper, we design a geometric reward function that rewards the agent if the vector defined by its forearm is aligned with the vector connecting its end-effector and the target as seen on Figure \ref{fig:angles}. 
\begin{equation}
\hat{\theta} = 1- \frac{\angle (\overrightarrow{V_{HT}}, \overrightarrow{V_{EH}})}{\pi}
   \quad\text{,}\quad 
r_t^{Pt}=\frac{e^{\hat{\theta}}-1}{e}
\end{equation}


\subsection{Results}

We compare the models on a separate set of 12 target positions outside of the training set (AMP training has seen these target positions but not the motions directly, so the comparison remains valid). To quantify overall pointing accuracy, we calculate reward statistics over the 12 pointing sequences for each model as shown in Table \ref{tab:model-r}. \textbf{AMP-fr} obtains highest $\mathbf{r_\text{max}}$, showing that it has learned to point more accurate than other models. \textbf{DM-32} has the highest pointing accuracy out of all DM models, which could be due to its higher number of training data. 

\begin{figure}[!t]
  \centering
  \vspace{0.5em}
 
  \includegraphics[width=1.0\linewidth]{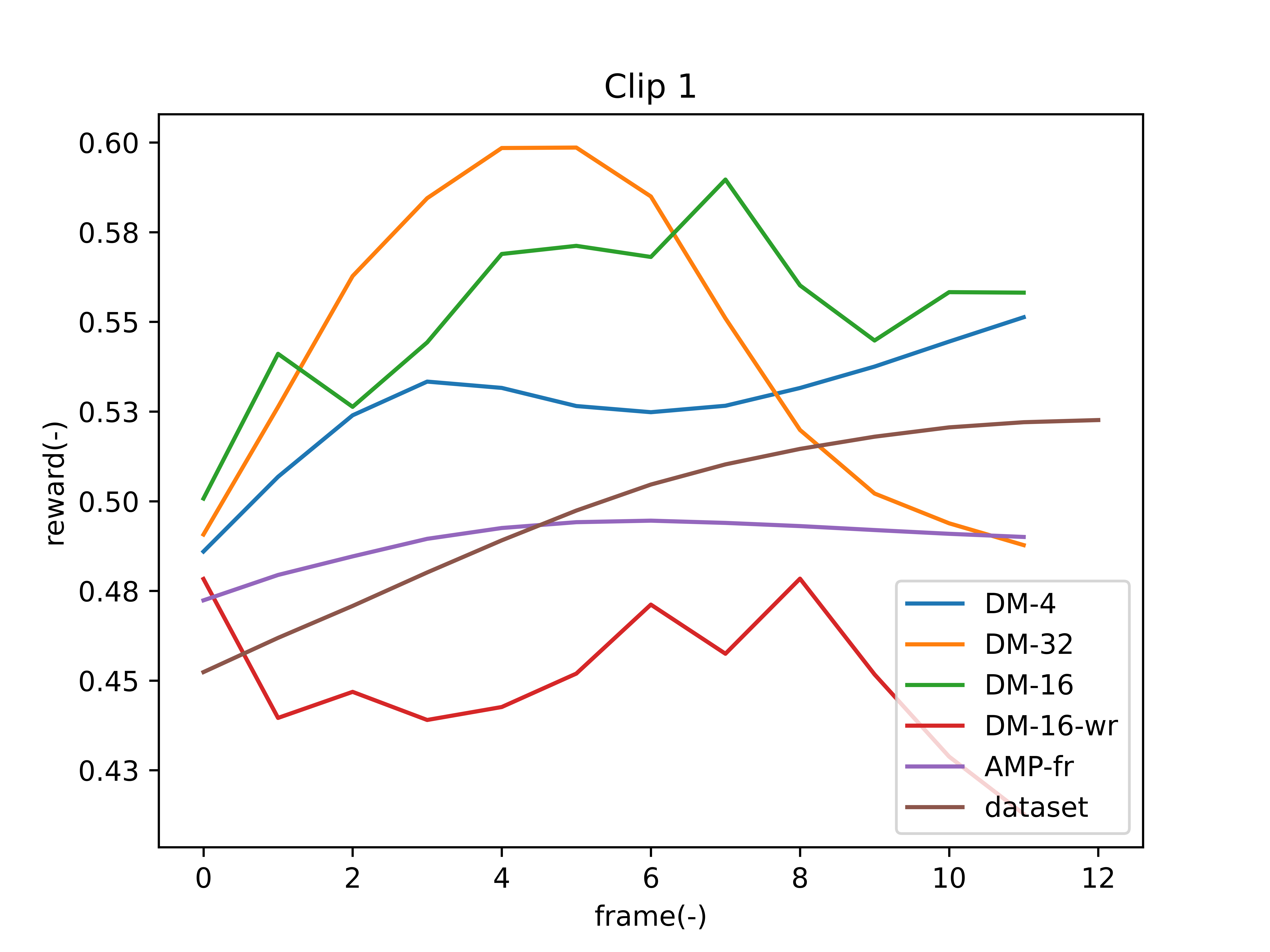}
  \\

    \includegraphics[width=1.0\linewidth]{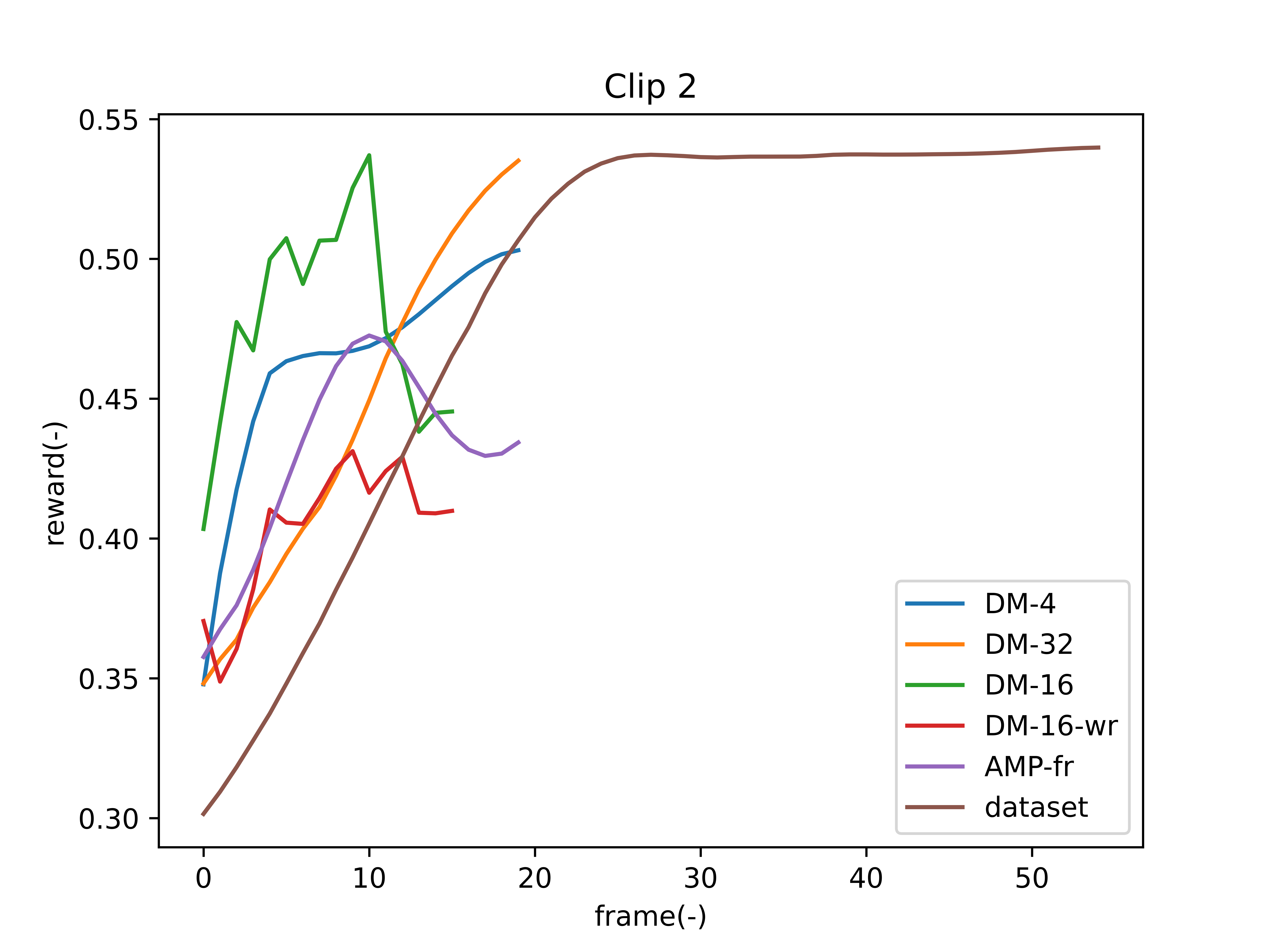}
  \caption{Reward comparison for different settings}
  \label{fig:reward_plots}
\end{figure}

We plot the angle rewards obtained by the models (along with the ground-truth motion) when starting at the same idle pose and given the same target position in Figure \ref{fig:reward_plots}. Two sample clips are shown. The horizon of frames ends at the highest ground-truth motion reward. \textbf{DM-32} in Clip 1 is an example of a good model pointing behavior, where both the raising and retraction motions are smooth as seen by the smoothness of the reward curves. The pointing is also accurate as the highest reward at the peak of bell curve is actually higher than ground-truth motion. The plots also show that several models make undesirable jerky, vibration-like motion towards the pointing target. 

\begin{table}[htbp]
\caption{Pointing rewards obtained by models}
\begin{center}
\begin{tabular}{|c|c|c|c|}
\hline
\
\textbf{} & $\mathbf{r_\text{max}}$ & $\mathbf{r_\text{min}}$ & $\mathbf{r_\text{mean}}$ \\
\hline
\textbf{DM-4} &  $0.53\pm0.08$ & $0.34\pm0.09$ & $0.46\pm0.07$  \\
\hline
\textbf{DM-16} &  $0.52 \pm 0.08$ & $0.37 \pm0.06$ & $0.46 \pm 0.07$   \\
\hline
\textbf{DM-16-wr} &  $0.44\pm0.08$ & $0.34\pm0.07$ & $0.39\pm0.07$    \\
\hline
\textbf{DM-32} & $0.55\pm0.09$ &  $0.34\pm0.07$  &  $0.48\pm0.08$   \\
\hline
\textbf{AMP-fr} &  $0.58 \pm0.05$ & $0.35 \pm 0.05$  & $0.47 \pm0.04$    \\
\hline
\multicolumn{4}{l}{$^{\mathrm{a}}$Each metric is first calculated per test motion clip (12 in total) and}\\
\multicolumn{4}{l}{then calculated summary statistics which are shown here.}
\end{tabular}
\label{tab:model-r}
\end{center}
\end{table}

\begin{table}[htbp]
\caption{Motion smoothness comparison}
\begin{center}
\begin{tabular}{|c|c|c|c|}
\hline
\
\textbf{} & \textbf{$\mathbf{vel_r}$ (r/t)}& \textbf{$\mathbf{acc_r}$ (r/t)} & \textbf{$\mathbf{jerk_r}$ (r/t)} \\
\hline
\textbf{DM-4} &  $0.33\pm0.26$ & $9.61\pm10.69$ & $739.11\pm940.16$  \\
\hline
\textbf{DM-16} &  $1.11\pm0.28$ & $84.74\pm19.06$ &  $9328.20\pm2349.19$    \\
\hline
\textbf{DM-16-wr} &  $0.94\pm0.16$ &  $81.19\pm17.74$  &  $8615.28\pm2449.04$    \\
\hline
\textbf{DM-32} &  $0.35\pm0.25$ & $8.87\pm8.85$ & $592.48\pm549.79$    \\
\hline
\textbf{AMP-fr} &  $0.47\pm0.20$ & $7.72\pm3.06$  & $350.85\pm130.22$    \\
\hline
\multicolumn{4}{l}{$^{\mathrm{a}}$Lower measures in all three metrics indicate smoother motion.}
\end{tabular}
\label{tab:model-smooth}
\end{center}
\end{table}

To further quantify difference in motion smoothness, we calculate the rate of change of angle reward r ($\mathbf{vel_r}$), acceleration ($\mathbf{acc_r}$), and jerk ($\mathbf{jerk_r}$). The results are shown in Table \ref{tab:model-smooth}. \textbf{AMP-fr} obtains lowest measures in all three categories indicating that its generated motions are smoother than other models. This is consistent with our visual inspection. \textbf{DM-16} obtains highest measures in all three categories, which is also consistent with our observation that it is indeed unstable and constantly makes jerky motions in pointing arm towards the target. These results suggest that AMP is advantageous in obtaining smooth motion and should be tested more extensively in future studies.

\section{Discussion}
This paper presents initial results of learning pointing skills in a physically simulated agent using reinforcement learning.  
In future studies, our main focuses are incorporating additional context in the form of task reward and re-evaluating how the pointing task is defined, which includes modifying the data collection procedure, and testing different inputs to the learning system and designing more appropriate reward functions. We also want to incorporate motion style using a custom style reward function.

The current setup aimed at reproducing the dynamics of pointing movements, which were recorded in a rigid setting. We  plan to  design  an experimental setup, which reflects everyday communication scenarios more closely (e.g. target selection in games). On the algorithmic side, we also plan to extend the system to accept observational data (ego-centric images) and speech as input. Pointing gestures are most often accompanied by verbal communication, which can provide context and can help resolving  disambiguities in the interlocutor's side in understanding the pointing gesture. Ego-centric images would allow us to learn hand-eye-coordination in the pointing task. 

We also plan to re-evaluate the reward function design, especially how we can achieve context-aware pointing by incorporating additional context in the form of reward functions. Achieving the intended objective in reinforcement learning is largely determined by devising an appropriate reward function, however this is often a challenging task. In this paper we took a reward function based on simple geometrical relations between the target and the agent's end-effector. However, there are no clear geometric rules for guiding the production and interpretation of pointing gestures even in human-human communication and people often fail to determine the target location from pointing gestures \cite{Herbort2018HowTP}, a finding also confirmed in analysis on our motion-captured pointing gesture dataset as the alignment angle between actor's arm and pointed target is not near 0. But humans can often refer to the object accurately with pointing in the interaction context. To this end we believe that, since pointing is a communicative signal, its success should also be measured by the recognition of the intended target. Thus adding context-aware recognition of pointed object to the learning loop could be key to a successful system.  In future studies we also intend to make use of user studies to evaluate the generated pointing behavior in context. These evaluations could include disambiguation from closely situated targets.

\balance
\bibliography{references}
\bibliographystyle{plain}

\end{document}